# Sparse Auto-Regressive: Robust Estimation of AR Parameters

Mohsen Joneidi

***Abstract:*** *In this paper a new approach for regression of time series using their own samples is presented. This is a celebrated problem known as Auto-Regression. Dealing with outlier or missed samples in a time series makes the problem of estimation difficult, so it should be robust against them. Moreover for coding purposes I will show that it is desired that the residual of auto-regression be sparse. To this aim, I first assume a multivariate Gaussian prior on the residual and then I obtain the estimation. Two simple simulations have been done on spectrum estimation and speech coding.*

## 1. Introduction

For more than 3 decades Autoregressive (AR) has been considered as an important model in signal processing. This model is exploited in many applications such as time-series analysis, speech processing and spectral estimation. In this model each sample of a signal is represented in terms of a linear combination of the other samples of the signal. The backward equation is:

$$\begin{bmatrix} x_{p+1} \\ \vdots \\ x_N \end{bmatrix} = \begin{bmatrix} x_p & \cdots & x_1 \\ \vdots & \ddots & \vdots \\ x_{N-1} & \cdots & x_{N-p} \end{bmatrix} \begin{bmatrix} a_1 \\ \vdots \\ a_p \end{bmatrix} + \begin{bmatrix} r_{p+1} \\ \vdots \\ r_N \end{bmatrix} \quad (1)$$

where, $x \in \mathrm{R}^N$ is a signal and $a \in \mathrm{R}^p$ is the corresponding autoregressive coefficients. $r \in \mathrm{R}^N$ is the residual error. Let denote the above equation in the following matrix form:

$$x = Xa + r \quad (2)$$

This problem is just a regression; a traditional problem is solved by minimizing the least squares estimation. The MSE minimization problem and its solution are:

$$a = \min_{a} \|r\|_2 \ st: r = x - Xa \quad (3)$$

$$a = (X^T X)^{-1} X^T x \quad (4)$$

If one assume Gaussian prior for the residual this solution is also MAP estimation and minimum entropy estimation. But Gaussian prior results in sensitivity to outlier and missing samples. Furthermore, many real signals does not fit with the Gaussian distribution because these signals are distributed with tails decaying slower than Gaussian and have peak taper than Gaussian (like speech signals). Gaussian assumption on the residual implies that the observations must to be Gaussian too, but neither the observations nor the residuals are Gaussian. If a presented sample deviates from the mean of the Gaussian distribution, this observation has an adverse effect on the estimations. To address this problem, [1] suggest imposing the mixture of Gaussians distributions for the residual. Parameters of this model need to be learned, so appropriate learning is an important problem that increases the complexity of the method. Another approach assumes the long tailed distributions, [2] exploited student-t distribution. In this paper this approach is exploited.

Recently sparse methods and low-rank representations [14,15] has attracted a lot of attentions and they have been exploited in many applications such as image denoising [12], voice detection [13], image classification and segmentation [10,11].

Moreover it is desired to make autoregressive model robust against missing data. [3] presented the application of autoregressive analysis in missing data problems. In [4] an algorithm was suggested for estimation of AR parameters in missing data situation. From statistical point of view, missing a data causes an outlier in the residual signal, thus the prior on the residual distribution must adopt outliers. Using long tailed distributions may be suitable to adoption in the outlier scenario. In this paper I propose to use multivariate Gaussian distribution as the prior on the residual. By this assumption the least squares regression problem in (2) converts to the sparse regression. An approximation of the sparse regression is least absolutes regression that [5] exploited this regression for robust AR parameters estimation. Least absolutes regression is a MAP estimator by Laplace distribution prior on the residual that it has longer tail in contrast with Gaussian.

## 2. Multivariate Gaussian for the residual distribution

In this section the multivariate Gaussian distribution will be assumed as the prior distribution for the residual of the regression:

$$p(r|W) = \frac{1}{(\sqrt{2\pi})^N |W|} \exp(-r^T W r) \quad (5)$$

The ML estimation problem for W is:

$$W_{ML} = \max_W p(r|W) \quad (6)$$

Let me first obtain an optimum W:

$$L(W) = \ln(p(r|W)) = r^T W r - \ln(|W|) + C$$

$$\frac{\partial L(W)}{\partial W} = 0 \rightarrow W_{ii} = \frac{1}{r_i^2}$$

$$W_{ii} = \frac{1}{\varepsilon + r_i^2} \rightarrow p(r) \propto \exp(-\|r\|_0)$$

where, $\varepsilon$ is a small positive and is introduced just for avoiding the division by zero. $\|r\|_0$ denotes the zero norm of a vector which is defined as the number of none-zero elements of it. Continuing with this distribution, the ML estimation of autoregressive coefficients is as follows:

$$a = \min_a \|r\|_0 \; st: r = x - Xa \quad (7)$$

The constraint of this problem is an over-determined system of linear equations that all the equations cannot be satisfied. [6] has named this system of equations the *"robust sensing"* problem. In robust sensing, the goal is to find vector $a$ which maximizes the number of equations. $x_i = X_i a$ is the i'th equation. Ref. [6] has showed that robust sensing is an NP-Hard problem and proved that the solution of robust sensing in a certain conditions equal to:

$$a = \min_a \|r\|_1 \; st: r = x - Xa \quad (8)$$

Problem (8) equals to the MAP estimation of $a$ by Laplace prior on the residual. Equivalence of (7) and (8) is not a surprise result because Laplace distribution agrees with sparse signals.

Problem (7) is robust to outliers and gross errors but this equation is very sensitive to a Gaussian residual because Gaussian assumption on the residual does not allow the residual to be exactly zero. Thus, only p equations can be satisfied simultaneously. To handle this problem, I suggest the following problem:

$$a = \min_a \sum_{i=1}^{N-p} \frac{r_i^2}{\rho + r_i^2} \; st: r = x - Xa \quad (9)$$

If $\rho$ tends to zero, this problem equals to (7). $\rho$ is a constant depends on the Gaussian variance ($\sigma$). Since 95% samples of a Gaussian process deviates less than $2\sigma$ from the mean thus $\rho = (2\sigma)^2$ seems good. By this $\rho$ the residuals less than $2\sigma$ approximately penalized by norm 2, residuals greater than $2\sigma$ approximately penalized by norm 0. This problem similar to (7) is also none convex.

To solve it I exploit Graduated non-convexity (GNC) technique [7] that will be described in Section 4.

## 3. Relation to sparse representation

To understand how sparse residual is related to the sparse representation in terms of over complete bases, I re-write (8) in the following form:

$$a = \min_a \|x - Xa - r\|_2 + \lambda \|r\|_1 \quad (10)$$

Defining $D = [X \; I]$:

$$a = \min_a \left\| x - D \begin{bmatrix} a \\ r \end{bmatrix} \right\|_2 + \lambda \|r\|_1 \quad (11)$$

$$z = \min_z \|x - Dz\|_2 + \sum_i \lambda_i |z_i| \quad (12)$$

(12) is a sparse decomposition problem with group sparse constraint. Making sparse the coefficients that corresponds to the identity matrix is equivalent to making sparse the residual of a regression problem. It's possible to exploit some other matrices instead of identity matrix that better makes the residual sparse. But the approach of this paper is not this idea. If computational burden does not have any care, research from this view will be more successful because learning appropriate dictionary for different components of signals provide suitable representation domain for processing. For example if the residual of our problem is band limited, then we can use DCT or FFT matrix instead of identity matrix to have more sparse residual.

## 4. Optimization

Problem (9) is a kind of M-estimator [8]. A well-known method to solve this type of problems is Iterative Re-weighted Least Squares (IRLS) [9]. Figure 1 depicts the algorithm for solving this problem.

---

Input: a time series of data ($x_t$), regression order (p)
Output: coefficients of auto-regression

1- Extract x and X from $x_t$ according to (1)
2- Initialization W
3- Loop (until convergence)

$$a = (X^T W X)^{-1} X^T W x$$

$$r = x - Xa$$

$$W = diag(\frac{1}{\rho + r_i^2})$$

---

Figure 1: The Optimization algorithm for the proposed regression

For small $\rho$ values, the whole problem is extremely non-convex. Thus it is very probable that the algorithm may be trapped in a local optimum. To reduce this probability the following criterion for re-weighting has been used:

$$W = diag(\frac{1}{\rho + |r_i|^p}) \quad (13)$$

Where, p is an increasing scalar tends to 2 as the iterations tend to infinity which is inspired by the idea of GNC.

## 5. Application to time series with missing data

Problems based on minimization of MSE are very sensitive to missing data, because missed data probably have large errors and their squared errors have enormous effect, while M-estimators like the proposed estimation are able to reduce the adverse effect of gross errors. In this section I just show an intuitive experience on the applicability of the proposed estimation by an example. Figure 2 shows a synthetic time series with 64 samples that 25% of their samples has been lost.

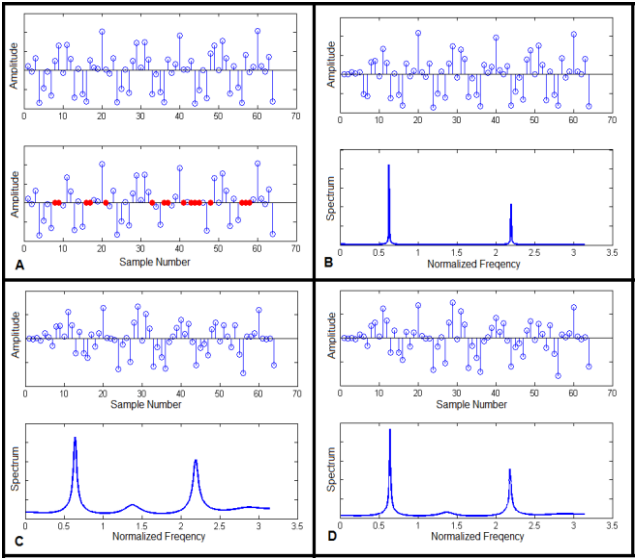

Figure 2: A simple simulation in situation with missing samples

Figure 2-A shows the original series and the series with missed samples. In figure 2-B the estimated time series and its spectrum using original time series has been depicted. Figure 2-C shows the estimated time series and its spectrum using missing time series have been calculated by solving yule-walker equations. Figure 2-D is similar to 2-C but using the proposed algorithm. As it can be seen, the estimated spectrum resolves the peaks of the spectrum more accurately.

## 6. Application to speech coding

As described previously, the problem (8) differs from traditional AR in the sense of residual distribution. In the proposed problem the residual has Laplace distribution. Let me compare the entropy of Gaussian and Laplace distributions. In this section assume that $p_G(r)$ and $p_L(r)$ are distributions of two residuals that both of them are corresponding to the regression of a signal. In the first residual, norm 2 of residuals is minimized and in the second one norm 1.

$$p_G(r) = \frac{1}{\sqrt{2\pi}\sigma_G} \exp(-\frac{r^2}{2\sigma_G^2}) \quad (14)$$

$$p_L(r) = \frac{1}{\sqrt{2}\sigma_L} \exp(-\frac{\sqrt{2}|r|}{\sigma_L}) \quad (15)$$

$$H_G(\sigma_G) = E[-\ln(p_G(r))] = \ln(\sqrt{2\pi e}\sigma_G) \quad (16)$$

$$H_L(\sigma_L) = E[-\ln(p_L(r))] = \ln(\sqrt{2}e\sigma_L) \quad (17)$$

$$\Delta H = H_G(\sigma_G) - H_L(\sigma_L) = \ln(\sqrt{\frac{\pi}{e}}\frac{\sigma_G}{\sigma_L}) \quad (18)$$

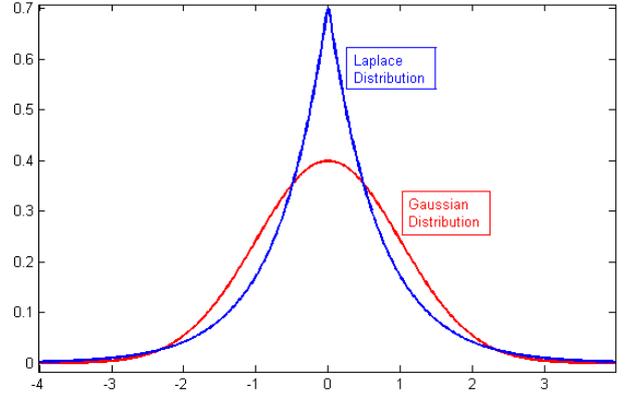

Figure 3: Comparison of Gaussian and Laplace distributions

$\Delta H$ shows the entropy of the difference of two Gaussian and Laplace sources. Figure 3 depicts both distributions with the same variance. The peak of Laplace distribution is sharper than Gaussian. In other words, the number of samples of a Laplace source around zero is more than Gaussian. But it is obvious that the variance of residual is minimized if Gaussian prior is assumed. Although many samples of a Laplace source are around zero, but few large samples is sufficient to make the variance large. If entropies of two Gaussian and Laplace sources are equal, the Gaussian source has less variance. Thus according to (18) coding of the Laplace residual can not be considered for lossless speech compression because the variance of Laplace residual may become large due to some large residuals. But a framework for a lossy coding scheme can be designed. If the samples of the Laplace residual corresponding to large errors be limited to a certain bound, variance of the residual decreases tremendously. Assume this bound is selected so that to satisfy:

$$\frac{\sigma_G}{\sigma_L} = L$$

$$\Delta H = \ln(\sqrt{\frac{\pi}{e}}L) \quad (19)$$

Since clipping a Laplace samples to a bound makes the peaks of the PDF to go on the borders, thus indeed the delta entropy is more than the equation (16). In the following I present an experiment on speech coding. For the simulation, 100 sentences from TIMIT database has been used. In this experiment, at first the signals are converted into some sets of non-overlapping frames and then auto-regression is applied in each frame. Afterward the residual for each frame is quantized to $2^{16}$ levels. Let me introduce the parameter K for the simulations that is equal to the fraction of maximum values of residuals for a signal. By this parameter the bound for the projection of large residuals will be determined. K is inversely related to L in equation (16). Samples which are larger than $K.\max(r)$ are projected on $K.\max(r)$, and those which are less than $K.\min(r)$ are projected on $K.\min(r)$. Figure 4-A shows the average SNR versus 1/K. for K=1 there is only quantization noise. But by decreasing K, since some samples are clipped (corresponds to large residuals) SNR decrease rapidly. In this figure SNR refers to the difference of the clipped and quantized samples with the original residual samples. Figure 4-B shows the average entropy versus 1/K. As expected entropy has direct relation to SNR.

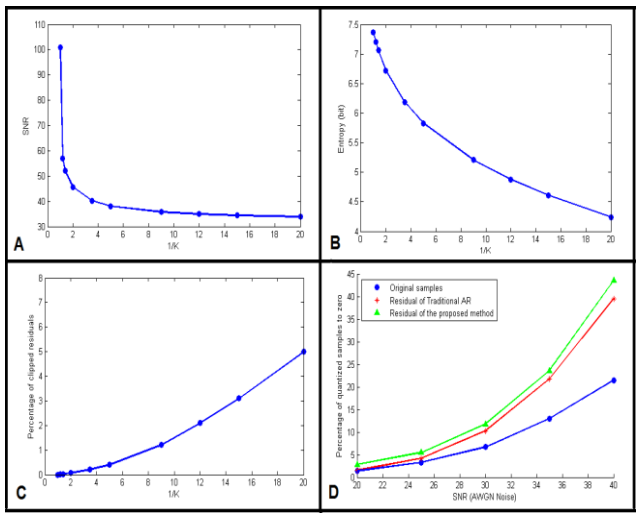

Figure 4: Application of proposed Auto-regression in speech coding

Figure 4-C shows the percentage of the number of residual samples that are clipped versus 1/K. for example if we put K=0.2 only near 0.5% of samples are clipped but the entropy decrease to near 21%. This property does not exist for the residual obtained by Gaussian prior for example if we clip the Gaussian residuals with K=0.2 near 0.9% of samples must be clipped for saving only %15 in entropy. Another advantage of Laplace distribution over Gaussian is concentration of residual samples around zero. Figure 4-D compares the percentage of residual samples that are quantized to zero level versus SNR of additive noise to the original signal. As expected, by AR coding the percentage of small residuals will increase in comparision with the original signal. It can be seen that AR with sparse constraint increases the number of samples around zero.

Note that SNR in the last figure is different from SNR in figure 3-A.

## 7. Conclusion

Sparse domains in signal processing make interpretations and designs simpler. As was seen in this paper, sparse residual for Auto-regression parameter estimation brings suitable properties for the residual signal. Only two applications of this approach has been studied but this robust estimation can be applied in many fields in which auto-regressive is an appropriate model.

**References**

[1] Roberts, S.J.; Penny, W.D.; , "Variational Bayes for generalized autoregressive models," Signal Processing, IEEE Transactions on , vol.50, no.9, pp. 2245- 2257, Sep 2002

[2] Christmas, J.; Everson, R.; , "Robust Autoregression: Student-t Innovations Using Variational Bayes," Signal Processing, IEEE Transactions on , vol.59, no.1, pp.48-57, Jan. 2011

[3] Lee, T.C.M.; Zhengyuan Zhu; , "Nonparametric spectral density estimation with missing observations," Acoustics, Speech and Signal Processing, 2009. ICASSP 2009. IEEE International Conference on , vol., no., pp.3041-3044, 19-24 April 2009

[4] Broersen, P.M.T.; de Waele, S.; Bos, R.; , "Estimation of autoregressive spectra with randomly missing data," Instrumentation and Measurement Technology Conference, 2003. IMTC '03. Proceedings of the 20th IEEE , vol.2, no., pp. 1154- 1159 vol.2, 20-22 May 2003

[5] Youshen Xia; Kamel, M.S.; , "A Generalized Least Absolute Deviation Method for Parameter Estimation of Autoregressive Signals," Neural Networks, IEEE Transactions on , vol.19, no.1, pp.107-118, Jan. 2008

[6] Kekatos, V.; Giannakis, G.B.; , "From Sparse Signals to Sparse Residuals for Robust Sensing," Signal Processing, IEEE Transactions on , vol.59, no.7, pp.3355-3368, July 2011

[7] Blake, Andrew; Zisserman, Andrew (1987). Visual Reconstruction. MIT Press

[8] P. J. Huber and E. M. Ronchetti, *Robust Statistics*. New York:Wiley, 2009

[9] Basu, A.; Paliwal, K.K.; , "Robust M-estimates and generalized M-estimates for autoregressive parameter estimation," TENCON '89. Fourth IEEE Region 10 International Conference , vol., no., pp.355-358, 22-24 Nov 1989

[10] Minaee, Shervin and Abdolrashidi, AmirAli and Wang, Yao "Iris Recognition Using Scattering Transform and Textural Features" arXiv preprint arXiv:1507.02177

[11] Minaee, Shervin and Wang, Yao "Screen Content Image Segmentation Using Least Absolute Deviation Fitting " IEEE Internation Conference on Image Processing 2015.

[12] Mohsen Joneidi, Mostafa Sadeghi, Mojtaba Sahraee-Ardakan, Massoud Babaie-Zadeh, Christian Jutten "A study on clustering-based image denoising: From global clustering to local grouping" Signal Processing Conference (EUSIPCO), 2014 Proceedings of the 22nd European.

[13] Parvin Ahmadi, Mohsen Joneidi "A new method for voice activity detection based on sparse representation" Image and Signal Processing (CISP), 2014 7th International Congress on.

[14] Mostafa Rahmani, George Atia "Randomized Robust Subspace Recovery for High Dimensional Data Matrices" available online: http://arxiv.org/abs/1505.05901

[15] Mostafa Rahmani, George Atia "High Dimensional Low Rank plus Sparse Matrix Decomposition" http://arxiv.org/abs/1502.00182